\documentclass[letterpaper]{article}
\usepackage{aaai24}
\usepackage{times}  
\usepackage{helvet}  
\usepackage{courier}  
\usepackage[hyphens]{url}  
\usepackage{graphicx} 
\usepackage{natbib}  
\usepackage{caption} 
\usepackage{algorithm}
\usepackage{algorithmic}

\makeatletter

\newcommand{\Rmnum}[1]{\expandafter\@slowromancap\romannumeral #1@}
\makeatother
\usepackage{newfloat}
\usepackage{listings}
\DeclareCaptionStyle{ruled}{labelfont=normalfont,labelsep=colon,strut=off} 
\floatstyle{ruled}
\newfloat{listing}{tb}{lst}{}
\floatname{listing}{Listing}
\usepackage{booktabs}
\usepackage{pifont}
\usepackage{amsmath}
\usepackage{amsfonts}
\usepackage{adjustbox}
\usepackage{multirow}
\usepackage{array}
\usepackage{color}
\usepackage{url}

\title{Open-Vocabulary Video Relation Extraction}
\author{
Wentao Tian, Zheng Wang, Yuqian Fu, Jingjing Chen, Lechao Cheng
}
\author {
    Wentao Tian,\textsuperscript{\rm 1} 
    Zheng Wang,\textsuperscript{\rm 2} \equalcontrib 
    Yuqian Fu,\textsuperscript{\rm 1} 
    Jingjing Chen,\textsuperscript{\rm 1}\equalcontrib 
    Lechao Cheng\textsuperscript{\rm 3}
}
\affiliations {
    \textsuperscript{\rm 1} Shanghai Key Lab of Intell. Info. Processing, School of CS,
Fudan University\\
    \textsuperscript{\rm 2} College of Computer Science and Technology, Zhejiang University of Technology\\
    \textsuperscript{\rm 3} Zhejiang Lab\\
    \{wttian22@m., fuyq20@, chenjingjing@\}fudan.edu.cn, zhengwang@zjut.edu.cn, chenlc@zhejianglab.com
}
\begin{document}
\maketitle
\begin{abstract}
A comprehensive understanding of videos is inseparable from describing the action with its contextual action-object interactions. However, many current video understanding tasks prioritize general action classification and overlook the actors and relationships that shape the nature of the action, resulting in a superficial understanding of the action. 
Motivated by this, we introduce \textbf{O}pen-vocabulary  \textbf{V}ideo \textbf{R}elation \textbf{E}xtraction (\textbf{OVRE}), a novel task that views action understanding through the lens of action-centric relation triplets. OVRE focuses on pairwise relations that take part in the action and describes these relation triplets with natural languages. Moreover, we curate the Moments-OVRE dataset, which comprises 180K videos with action-centric relation triplets, sourced from a multi-label action classification dataset. With Moments-OVRE, we further propose a cross-modal mapping model to generate relation triplets as a sequence. Finally, we benchmark existing cross-modal generation models on the new task of OVRE. Our code and dataset are available at 
\url{https://github.com/Iriya99/OVRE}.
\end{abstract}

\section{Introduction}
Videos contain abundant semantic information, including action, actors (e.g. humans, animals, objects, and other entities), and relationships between actors.
To comprehend the dynamic and complex real-world situations depicted in videos, researchers have investigated a wide range of video comprehension tasks.
These endeavors allow for the transition of understanding video content from broad semantic concepts to more detailed ones. 
Despite the variations, all of these tasks converge on a pivotal aspect 
: extracting semantic information within the videos and constructing a higher-level representation to facilitate comprehension. 
Foundational tasks, such as action classification~\cite{kong2022human} and temporal action localization~\cite{9062498}, primarily center on recognizing broad-level actions within videos, yet they often overlook the specific scenario in which these actions unfold. Consequently, these tasks often struggle to offer a profound understanding of the context and the specific actors that are part of actions. In essence, they concentrate solely on deciphering ``what" action is transpiring and ``when" it takes place, omitting the ``who" and ``how" aspects.
On the other hand, video captioning~\cite{chen2019deep}, video grounding~\cite{chen2019semantic, wang2021visual}, and video-text retrieval~\cite{song2023relation} strive to encapsulate the videos' essence through textual descriptions by mapping them into a joint semantic space. Nevertheless, these textual descriptions often provide a detailed-level overview of the action context, lacking a nuanced comprehension of relations. 

\begin{figure}[t]
  \centering
  \resizebox{0.7\linewidth}{!}
  {\includegraphics{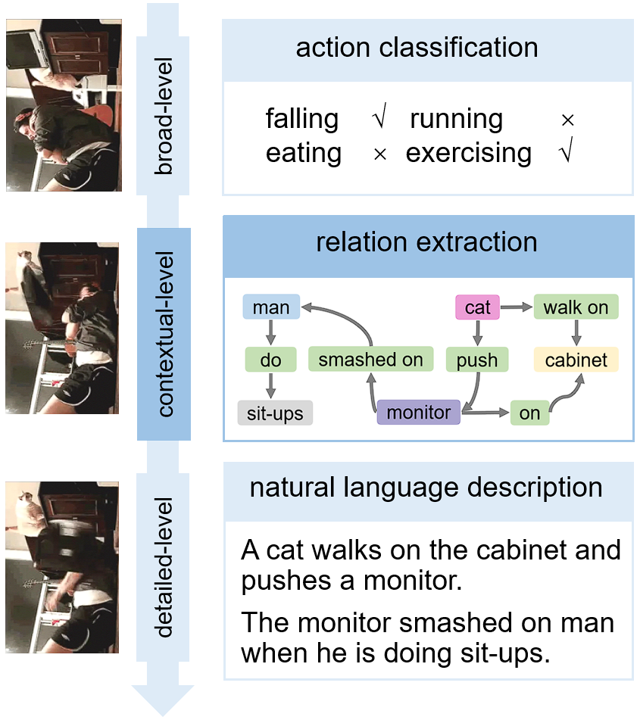}}
  \caption{Open-vocabulary  Video Relation Extraction enables a contextual-level comprehension of video content, bridging the gap between general action classification and precise language description.}
  \label{fig:motivation}
\end{figure}

\begin{figure}
  \centering
  \resizebox{\linewidth}{!}{\includegraphics{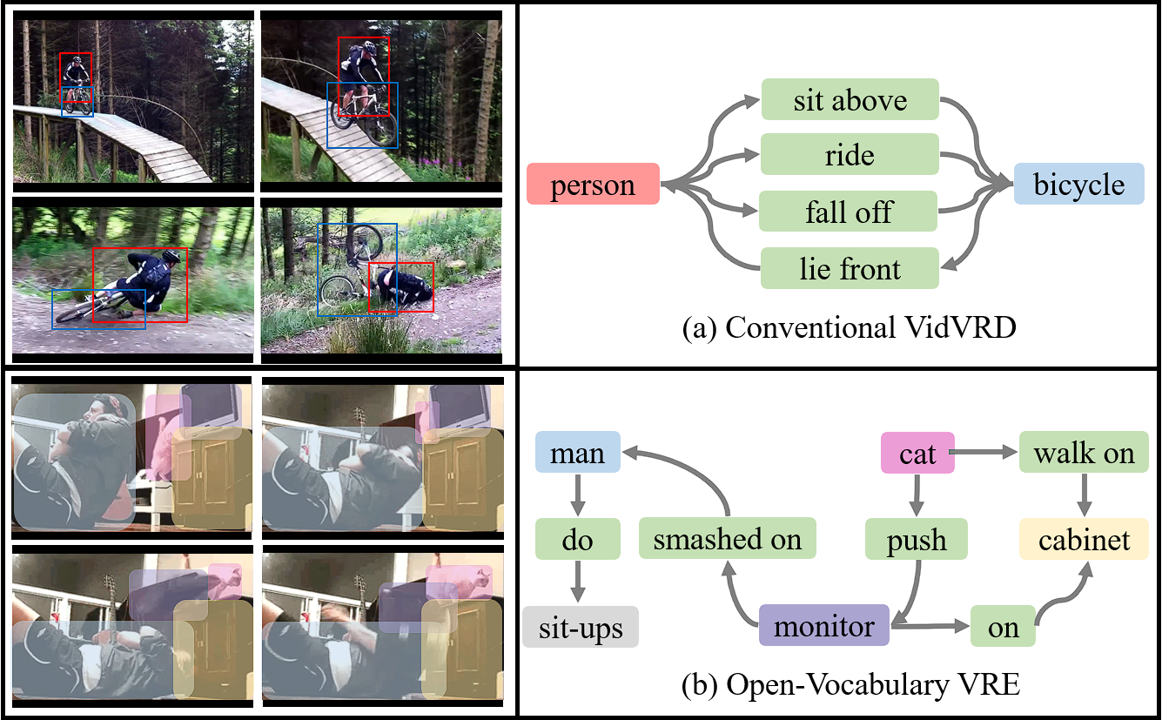}}
  \caption{Two \texttt{falling} videos depicted in VidVRD and OVRE diverse in (a) salient objects interconnected throughout frames are exhaustively annotated with diverse relations in VidVRD; (b) OVRE builds a relation graph with various actors and relations that closely related to the falling action.
  }
  \label{fig:compare_anno}
\end{figure}

To achieve a comprehensive comprehension of visual content, researchers have introduced Video Visual Relation Detection (VidVRD) tasks~\cite{Shang2017VideoVR,shang2019annotating} specifically designed for video analysis. VidVRD tasks are geared toward identifying objects and the relations between them within videos.
For instance, VidVRD~\cite{Shang2017VideoVR} employs available video object detection data to non-exclusively label various simplistic relationships between objects. As depicted in Figure~\ref{fig:compare_anno} (a), relationships like \texttt{touch} and \texttt{watch} are inherently \emph{object-centric} relationships with limited implications for the \texttt{falling} action.
Notably, Action Genome~\cite{genome} is distinct in its dedication to an \emph{action-centric} understanding of videos. It dissects actions into spatio-temporal scene graphs. However, it still grapples with a challenge shared by VidVRD – the constraint of a fixed vocabulary range. Consider the scenario shown in Figure~\ref{fig:compare_anno} (a) involving a bicyclist falling from a steep bridge. Due to the vocabulary limitations, the description fails to encompass the bridge and its relationship with the biker, which is an essential element in understanding the falling action.

To mitigate the aforementioned limitations of VidVRD, we introduce \textbf{O}pen-vocabulary \textbf{V}ideo \textbf{R}elation \textbf{E}xtraction (OVRE). OVRE extracts all \emph{action-centric} relation triplets from videos with language descriptions. Take Figure~\ref{fig:compare_anno} (b) as an example: OVRE dissects the \texttt{falling} action into multiple relation components, such as  \textless \texttt{cat}, \texttt{push}, \texttt{monitor} \textgreater and  \textless \texttt{monitor}, \texttt{smashed on}, \texttt{man}\textgreater.
Departing from VidVRD~\cite{Shang2017VideoVR} that confined objects and relations to limited categories, we harness the immense potential of large Language Models (LLMs) to articulate action-aware relations using natural language. In essence, we undertake a two-fold process: first, using CLIP visual encoder to encapsulate video semantics; second, mapping these semantics into the linguistic realm for generating relation triplets through a pre-trained LLM. Employing this straightforward baseline, we achieve the ability to articulate relation triplets with unconstrained vocabularies.

To study OVRE, we present Moments-OVRE, an extensive dataset encompassing over 180K diverse videos.
These videos are sourced from a subset of Multi-Moments in Time~\cite{9609554}(M-MiT), a multi-label video dataset with an average video duration of three seconds.
In Moments-OVRE, we annotate both actors and relations as relation triplets that are most relevant to the action labels without vocabulary limitations. Note that ``actor" does not refer only to human individuals but can also refer to objects, animals, or any entities that are actively involved in an action.
Moments-OVRE confers several advantages:
(\Rmnum{1}) Unrestricted Vocabulary Annotations: Moments-OVRE benefits from an expansive vocabulary encompassing diverse actors and relations, offering a more accurate portrayal of real-world scenarios.
(\Rmnum{2}) Emphasis on Action-Centric Annotations: Our annotations exclusively focus on relation triplets related to actions/events depicted in the video.
(\Rmnum{3}) Ample Dataset Scale: Moments-OVRE contains over 180K videos, establishing itself as the most extensive video relation extraction dataset known to date.


The main contributions are as follows:

\textbullet \ We introduce the novel Open-vocabulary Video Relation Extraction task, aiming to extract action-centric relations within videos. 

\textbullet \ To foster ongoing research, we curate Moments-OVRE, an expansive dataset for video action comprehension. Comprising over 180k videos, Moments-OVRE is enriched with comprehensive open-vocabulary relationship annotations.

\textbullet \ We build a simple pipeline designed for generating action-centric relation triplets from raw videos. Extensive experimentation is conducted to validate assorted design considerations employed within the pipeline.

\section{Related Works} 









\begin{table*}[t]
    \centering
    \caption{Representative video understanding datasets.}
    \begin{adjustbox}{width=\textwidth}
    \begin{tabular}{cccccccc}
        \toprule
        Dataset & Task & Annotation & \#Video & \#Subject/Object & \#Predicate & Avg. Time (s) & Source \\
        \midrule
        Kinetics400 & \multirow{2}{*}{Action Classification} & \multirow{2}{*}{Action Categories} & 306,245 & - & - & 10 & YouTube\\
        VideoLT &  &  & 256,218 & - & - & 192 & YouTube\\
        \midrule
         MSR-VTT & \multirow{2}{*}{Video Captioning / Retrieval} & \multirow{2}{*}{Textual Descriptions} & 7,180 & - & - & 20 & Web\\
         VATEX & & & 41,269 & - & - & 10 & YouTube\\
        \midrule
        VidVRD & \multirow{3}{*}{Video Visual Relation Detection} & \multirow{3}{*}{BBoxes, Relation Triplets} & 1,000 & 35 & 132 & 33 & ILVSRC2016-VID\\
        VidOR & & & 10,000 & 80 & 50 & 35 & YFCC100M\\
        Action Genome & & & 10,000 & 35 & 25 & 30 & Charades\\
        \midrule
        VidSitu & Video Semantic Role Labeling & Verbs, SRLs, Event Relations & 29,220 & Open & Open & 10 & Condensed-Movies\\
        \midrule
        \textbf{Moments-OVRE} & \textbf{Open-Vocabulary Video Relation Extraction} & \textbf{Relation Triplets} & \textbf{186,943} & \textbf{Open} & \textbf{Open} & \textbf{3} & \textbf{M-MiT}\\

        \bottomrule
        \\
    \end{tabular}
    \end{adjustbox}
    \vspace{-15px}
    \label{tab:dataset_comp}
\end{table*}

\textbf{Video understanding} has long been an active research area in computer vision, encompassing a wide range of diverse aspects.
Action recognition~\cite{Kay2017TheKH,8237884,monfortmoments} categorizes the video content into a predefined set of action classes, temporal action localization~\cite{Hei2015anet,liu2022fineaction} further emphasizes the specific time points at which these actions occur.
Tasks such as video captioning~\cite{chen2019deep}, video question answering~\cite{qian2023locate}, and video-text retrieval~\cite{song2021spatial} are dedicated to learning a detailed representation of semantic information.
In the spectrum of these tasks resides an intermediary domain referred to as relation understanding,  dedicated to capturing the diverse relations among objects. 
This contextual domain provides semantically rich intermediate representations for videos, crucial for comprehending interactions and behaviors depicted within them.
VidVRD~\cite{Shang2017VideoVR} and VidOR~\cite{shang2019annotating} are representative benchmarks for relation understanding, with finite object and predicate sets. 
Recently proposed task Open-VidVRD \cite{gao2023compositional} manually splits the finite object and predicate sets into base and novel categories and builds the novel prediction setting.
However, these existing VRD tasks focus on object-centric relation detection and struggle to bridge the gap between closed-label sets and natural languages, prompting us to propose video relationship extraction under open-vocabulary scenarios.
Moreover, we aim to identify informative triplets that portray various actions in the video.

\noindent \textbf{Contextual-level Visual Comprehension} has been mainly explored in the image domain, including scene graph generation~\cite{genome}, visual semantic role labeling~\cite{yatskar2016situation, sadhu2021visual}, and human-object interaction~\cite{li2022hake, gkioxari2018detecting}. These tasks have also been extended to the video domain, which requires information aggregation across frames.
Existing works often treat video relation detection as a multi-stage pipeline, including object tracking, tracklet pair generation, and relation classification. These works focus on improving relation classification by leveraging contextual knowledge~\cite{10.1145/3343031.3351058} or adding additional modules to enhance predicate representation~\cite{gao2021video} with generated tracklets.
Recent trends lean towards one-stage models like VRDFormer~\cite{zheng2022vrdformer}, employing various queries to integrate spatio-temporal information, facilitating tracklet pair generation and relation classification concurrently.
However, their scalability and generalization are hampered by datasets containing a restricted size of objects and relationships.
Video captioning, aimed at generating open-ended descriptions, typically uses an encoder-decoder architecture. Over time, this setup has advanced to more sophisticated models~\cite{zhang2019reconstruct,tang2021clip4caption,lin2022swinbert}. 
Recent strides in visual-language representation learning have led to models pre-trained on extensive paired data demonstrating promise in fine-tuning for various multi-modal tasks~\cite{wang2022git,xu2023mplug2,chen2023valor}.
Building on the progress in video captioning, we seek to leverage generative models for our OVRE task.

\noindent \textbf{Datasets for video understanding} have been introduced in support of multiple tasks over the past years.
Kinetics400 \cite{Kay2017TheKH} is a widely adopted category-balanced classification dataset. VideoLT~\cite{zhang2021videolt}, in opposition, follows a naturally long-tailed distribution for action recognition.
MSR-VTT \cite{7780940} is a widely adopted dataset for video captioning and video retrieval, it includes video clips from different domains, and each of them is annotated with approximately 20 natural sentences. Recently introduced VATEX~\cite{wang2019vatex} is a large-scale multilingual video description dataset.
VidVRD \cite{Shang2017VideoVR} and VidOR~\cite{shang2019annotating} densely annotate objects and the most basic relations.
Action Genome~\cite{genome} focuses on action understanding by decomposing them into different relationships and ignoring others that are irrelevant to actions. 
VidVRD datasets have restricted their objects and relations to a fixed label set and thus are far away from reflecting the diverse relations in the real world.
VidSRL~\cite{sadhu2021visual} presents the sole open-vocabulary relation understanding task, while it primarily focuses on events. In contrast, OVRE places greater emphasis on actions, which is a fundamental ability for event comprehension.
Detailed statistics for these datasets are listed in Table~\ref{tab:dataset_comp}.

\section{Video Relation Extraction Benchmark}
\subsubsection{Task Formulation}
Given a raw video input $V$, OVRE targets inferring its corresponding visual relationships $R$. Here, $V\in\mathbb{R}^{T\times H\times W\times C}$ denotes its frame sequence, where $T$, $H$, $W$, and $C$ represent the number of frames, width, height, and channels respectively. The relation triplets are represented as $R=\{{r_{1}},\cdots,{r_{K}}\}$, and each relation $r_{K}$ is structured as a triplet:  \textless \texttt{subject}, \texttt{relation}, \texttt{object}\textgreater.


Given that actions depicted in videos can encompass a wide spectrum of interactions, and distinct interactions might be common across various actions, our objective is to systematically extract all relationships associated with actions within the video context. 
Each $r_{i}$ consists of a sequence of words $w=\{w_{1},w_{2},...,w_{N}\}$. These individual $r_{i}$ are then concatenated to form a token sequence $R=w_{1},w_{2},...,w_{M}$, which serves as a representation of a sequence of relation triplets, collectively portraying all the action-centric relationships.
The training objective is to maximize the likelihood of generating $R$ given the video $V$:
$$
\underset{\theta}{\max} \log p_{\theta}(R|V),
$$
where $\theta$ denotes the model's trainable parameters.
Our key idea is to leverage the powerful capability of generative models to address the OVRE problem. 
Consequently, the revised training objective can be formulated as follows:
$$
\underset{\theta}{\max}\sum_{i=1}^{M}\log p_{\theta}(w_{i}|V,w_{i-1}).
$$
OVRE involves generating relationships using nature language, rendering common metrics wildly used VidVRD metrics such as Precision@K and Recall@K unsuitable for our task.
Inspired by the evaluation metrics in image captioning, we advocate for assessing the quality of generated triplets using Bleu, CIDEr, and METEOR scores.
For the Bleu metric, we only consider B@1, B@2, and B@3, as each triplet contains at least three words.
These metrics face challenges when applied directly to evaluating the overall relation triplets since the generated triplet sequence can be regarded as an unordered set, whose alignment with the ground truth triplets is unknown. 

\begin{figure}[t]
  \centering
  \resizebox{1.0\linewidth}{4cm}
  {\includegraphics{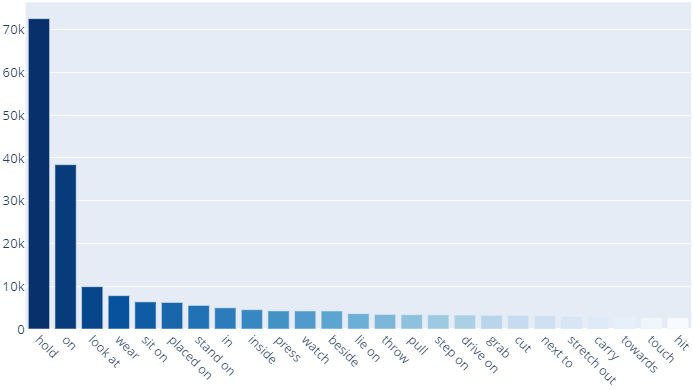}}
  \caption{Counts of top 25 relationships in Moments-OVRE.}
  \label{fig:freq}
\end{figure}

To match the generated triplet set with the ground truth, 
we first utilize a sentence embedding model SimCSE \cite{gao2021simcse} to quantify the text similarity between the generated and ground truth triplet set. 
Then we use these embeddings to acquire a similarity matrix $S$, where $S_{ij}$ denotes the cosine similarity score of generated triplet $\hat r_i$ and ground truth triplet $r_{j}$.
After obtaining $S$, we use the Hungarian algorithm to establish one-to-one matches between the triplets and apply the aforementioned metrics to these paired triplets.
Note that the number of generated triplets may differ from that of ground truth triplets. When fewer triplets are generated, unmatched ground truth triplets receive zero scores.

\section{Dataset Construction}
We present the Moments-OVRE dataset tailored for OVRE. Notably, Moments-OVRE boasts distinct attributes such as diverse video content, an extensive compilation of video-triplet pairs, annotations embracing open vocabulary, and a focus on action-centric relationships.
In this section, we will detail how we select representative videos, annotate relations, and devise data splits. Additionally, we thoroughly explore and analyze the statistical insights within the Moments-OVRE dataset.

\subsubsection{Video Selection}
Since our task focuses on action-centric relationships, several requirements are posed for the selection of videos. 
First, we prefer videos with multiple actions. Second, videos should contain rich information, involving different scenes, objects, and events in the real world.
We choose to annotate videos from Multi Moments in Time (M-MiT) \cite{9609554} due to its multi-label nature, which allows a more nuanced comprehension of actions within intricate contexts.
M-MiT offers several distinct advantages: 
(\Rmnum{1}) Unlike commonly utilized action datasets like Kinetics, M-MiT exhibits substantial intra-class diversity. This highlights the necessity of providing more intricate descriptions of relationships to capture variations between videos of the same action labels.
(\Rmnum{2}) A majority of M-MiT videos have a duration of only 3 seconds. This temporal brevity encourages annotators to focus primarily on portraying the action itself, as opposed to the temporal context which is more prevalent in longer videos, such as those in YouCook and Howto100M datasets.
Considering the long-tail distribution of action categories in M-MiT, we attempt to relatively balanced sampling videos from all classes. M-MiT has 292 action categories, we sample at least 660 videos per class and additionally select other random videos as some videos may be discarded during annotation. 
\begin{figure}[t]
  \centering
  \resizebox{0.85\linewidth}{!}{\includegraphics{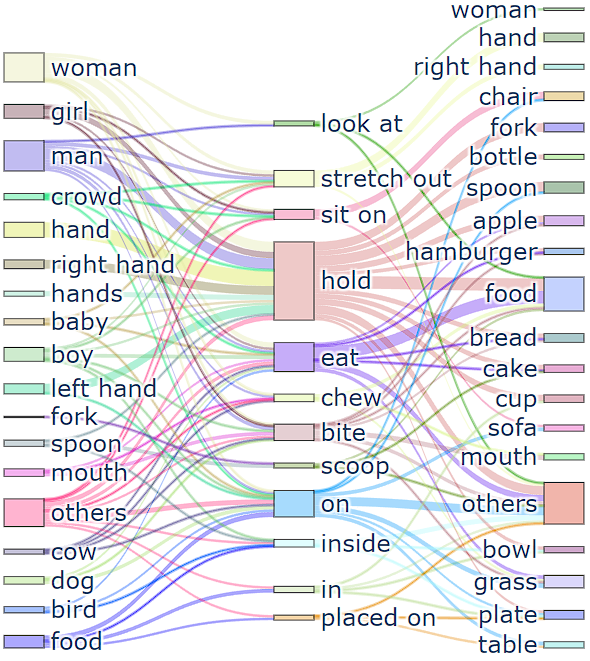}}
  \caption{A weighted bipartite mapping of the top 12 most frequent relations in \texttt{eating} videos.}
   \label{fig:sanki}
\end{figure}
\subsubsection{Annotation Pipeline}
We ask annotators to perform an action-centric relation triplet annotation, which follows these steps:
Given the potentially noisy nature of the large-scale M-MiT dataset, annotators first verify the correctness of the action labels. Videos with incorrect labeling are subsequently excluded.
To perform action-centric annotation, both the video and the corresponding action labels are presented to the annotators. They first identify and annotate all pertinent objects, and then articulate the associated relationships among these objects.
Annotators are instructed to provide descriptions for relation triplets solely when these relationships are relevant to the action labels.
To illustrate, refer to Figure~\ref{tab:dataset_comp} (b): the annotation \textless \texttt{monitor}, \texttt{smashed on}, \texttt{man} \textgreater is required, whereas the annotation \textless \texttt{guitar}, \texttt{placed on}, \texttt{ground} \textgreater is invalid, as the latter constitutes background information rather than being directly tied to the \texttt{falling} action.
Besides, we also manually review and correct the low-quality annotations.

\subsubsection{Dataset splits}
The data is partitioned into training and testing sets, resulting in 178,480 and 8,463 videos respectively.
To ensure maximum consistency with the original splitting, all videos are selected exclusively from their designated sets within the M-MiT collection.
\begin{figure*}[t]
  \centering
  \includegraphics[width=0.9\textwidth]{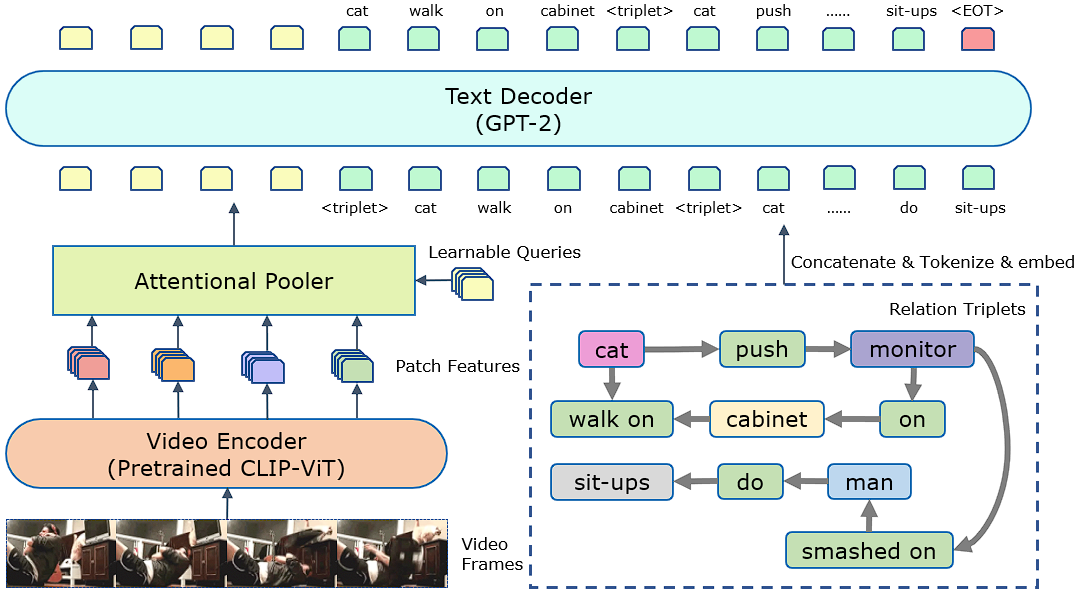}
  \caption{Overview of our model architecture, enabling simple relation generation with the powerful vision-language pre-trained model CLIP-ViT as the video encoder and the large language model GPT-2 as the text decoder.
  }
  \label{fig:arch}
\end{figure*}

\subsection{Dataset Analysis and Statistics}
Overall, Moments-OVRE offers open-vocabulary annotations for 186,943 videos, encompassing a sum of 399,576 relation triplets. 
In Figure~\ref{fig:freq}, we show the distribution of the top 25 most frequent relations in Moments-OVRE, which follows a natural long-tailed distribution. Furthermore, Figure~\ref{fig:sanki} shows frequently described relation triplets of \texttt{eating} videos. Note that \texttt{hold} is a prevalent visual relation shared by various kinds of videos, such as \texttt{painting}, and \texttt{drinking} videos. A clue from this observation is that merely relying on a single relation is insufficient to infer the action categories from the videos, and a comprehensive understanding requires recognition of combinations of diverse relations. 

\section{Method}
\subsubsection{Overview}
An illustration of our method is provided in Figure~\ref{fig:arch}. The overall framework mainly includes a Video Encoder, an Attentional Pooler, and a Text Decoder. 
The video is first transformed into a sequence of features via the video encoder.
These features are further condensed into prefix information by the attentional pooler, which is then fed into the text decoder for relation triplet generation.

\subsubsection{Video Encoder}
To extract visual features from a video, we leverage the visual encoder of a pre-trained CLIP \cite{radford2021learning} model.
Previous research has demonstrated its exceptional performance across various open-vocabulary recognition tasks \cite{ni2022expanding,tang2021clip4caption,luo2022clip4clip}. 
Our choice for visual feature extraction is CLIP-ViT, which utilizes ViT\cite{dosovitskiy2020image} to translate video frames into sequences of patches. These visual patches capture finer-grained information, making them an optimal fit for OVRE. However, the incorporation of the temporal dimension necessitates consideration of the resultant number of patches. Specifically, for a video $V\in \mathbb{R}^{T\times H\times W}$, this translates to $T\times\frac{HW}{P^{2}}$ patches.
\subsubsection{Prefix Strategies}
With the features from $V$, our focus shifts towards transmitting them into the textual domain for relation generation. A simple solution is to directly feed all patches into the decoder, while the excessive number of patches not only leads to additional computational costs but also introduces a lot of redundant information.
In light of this, we utilize an attentional pooler denoted as $F(\cdot)$, which employs a predetermined number of queries to extract meaningful features from all video patches:
$
q_1,...,q_m=F(p_1,...,p_n,q_1,...,q_m),
$
which aggregates spatial-temporal features into a more concise representation. 
\subsubsection{Attentional Pooler}
Drawing inspiration from the approach introduced in VideoCoCa \cite{yan2023videococa},
our framework incorporates an essential element known as the Attentional Pooler. This module takes learnable queries and patch features as its input, facilitating cross-attention mechanisms between them.
The optimization of both the parameters of the Attentional Pooler and the queries enables the queries to gradually refine their ability to extract significant relationships from the patches. This design proves well-suited to the demands of OVRE. Remarkably, even with a single-layer transformer, our model showcases its ability to generate a diverse set of meaningful visual relationships.

\subsubsection{Text Decoder}
We employ GPT-2 \cite{radford2019language} as our generation model.
Though relationships themselves are unordered, we observe that sorting them by Triplet Linearization \cite{cabot2021rebel} does not provide significant benefits in OVRE.
Accordingly, all relation triplets of a video are simply concatenated into a text sequence by a separation token.
Then, we map all the words into their corresponding tokens and pad them to the maximum length $M$ to obtain a sequence of embeddings.

\section{Experiments}
\subsection{Experiments Settings}
\noindent\textbf{Video and Text Preprocessing. } 
For each input video, we first resize it to 224 $\times$ 224 and then extract 16 frames using uniform sampling, resulting in 786 patches for each video. We utilize RandAugment \cite{Ekin2019randaugment} as our data augmentation strategy.
For the paired relation triplets, we concatenate the unordered relation triplets into a sequence using the special token \textless\texttt{triplet}\textgreater.

\noindent\textbf{Training Setting. }
We train the generation model using cross-entropy loss and employ teacher forcing to accelerate the training process.
All models are optimized using AdamW optimizer, with $\beta_{1}=0.9$, $\beta_{2}=0.999$, a batch size of 16, and weight decay of 1e-3.
The initial learning rate is set to 1e-6 for CLIP, 2e-5 for GPT-2, and 1e-3 for AttentionPooler.
We applied learning rate warm-up during the early 5\% training steps followed by cosine decay.
We trained the networks for 50 epochs on 8 Nvidia V100 GPUs and chose the model with the highest CIDEr score as the final model.

\begin{table}[t]
    \centering
    
    \begin{tabular}{cccccc}
        \toprule
        Method & B@1 & B@2 & B@3 & CIDEr & METEOR \\
        \midrule
        ClipCap & 29.75 & 16.32 & 9.48 & 125.45 &  19.25\\
        GIT & 35.19 & 20.12 & 11.90 & 155.38 & 23.06 \\
        \textbf{Ours} & \textbf{37.27} & \textbf{21.92} & \textbf{13.90} & \textbf{174.47} & \textbf{25.07} \\
        \bottomrule
        \\
    \end{tabular}
    \caption{Baseline comparison on Moments-OVRE.}
    \label{tab:main_res}
\end{table}
\begin{table}[t]
    \centering
    
    \begin{tabular}{cccccc}
        \toprule
        CLIP & GPT2 & CIDEr & METEOR \\
        \midrule
         \ding{55} & \ding{55} & 131.85 &  19.84\\
        \ding{55} & \ding{51} & 165.67 & 24.12 \\
       \ding{51} & \ding{51} & \textbf{174.47} & \textbf{25.07} \\
        \bottomrule
        \\
    \end{tabular}
    \caption{Study on vision and language model fine-tuning. The \ding{51} mark means fine-tuning the corresponding module.}
    \label{tab:abl_ft}
\end{table}
\begin{table}[t]
    \centering
    
    \begin{tabular}{ccc}
        \toprule
        Features & CIDEr & METEOR \\
        \midrule
        Region & 77.38 & 13.14 \\
        Frame & 115.85 &  18.11\\
        Patch & \textbf{165.67} & \textbf{24.12} \\
        \bottomrule
        \\
    \end{tabular}
    \caption{Comparisons with different visual features (w/o fine-tuning visual encoder).}
    \label{tab:abl_feat}
\end{table}

\subsection{Main Results}
\noindent\textbf{Baselines models. }
Previous VidVRD methods focused on predicting relationships over detected objects, which is essentially a classification task and thus cannot be applied to OVRE.
Therefore, we introduce several generative models as baseline models. 

\textbullet  \ ClipCap \cite{mokady2021clipcap} is an image captioning model that utilizes the same visual encoder and text decoder as we do.
It uses a mapping network to convert CLIP embeddings into GPT-2 prefixes.
To apply ClipCap for videos, we follow the most common strategy that treats each video frame as an individual image and then perform a mean pooling layer along the temporal dimension to obtain a global video representation. 

\textbullet  \ GIT \cite{wang2022git} stands as a vision-language generative model, demonstrating strong performance across numerous generation tasks. This achievement is attributed to its effective optimization of the language model loss during pre-training, involving a substantial collection of image-text pairs.
We directly fine-tune $\text{GIT}_{\text{B}}$ to generate relation triplets without making further modifications.





\noindent\textbf{Result and Analysis. }
We present our results on Moments-OVRE in Table~\ref{tab:main_res} and compare our approach with baseline methods trained under the same training settings. 
Our approach outperforms baseline generative methods, achieving a higher METEOR score (+6.22) than ClipCap and (+2.01) than GIT. 
We find that although GIT was pre-trained on 0.8B image-text pairs and achieved impressive performance on video captioning datasets, it did not perform as well as our approach on the OVRE task. This could be attributed to the fact that the image-text generative pre-training does not directly facilitate the understanding of fine-grained information such as relationships in videos.

\begin{figure}
  \centering
  \resizebox{0.85\linewidth}{!}{\includegraphics[width=0.4\textwidth]{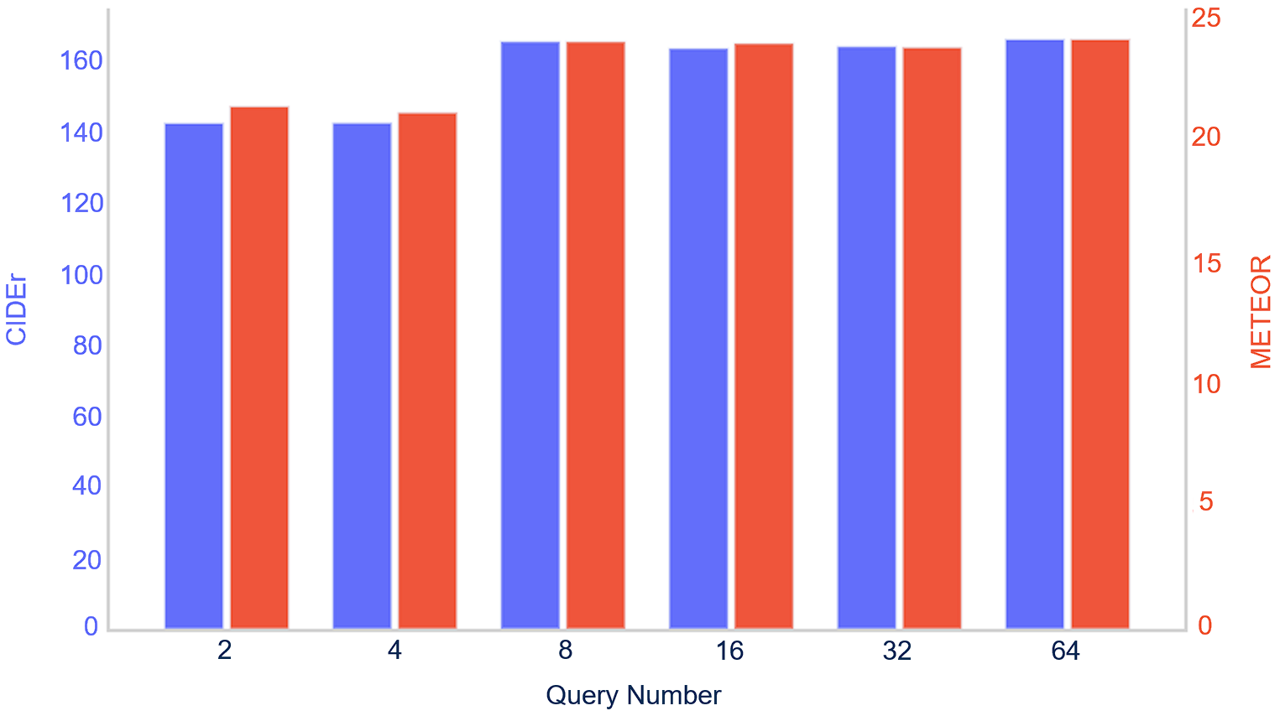}}
  \caption{Impact of the query numbers on the performance. For each query number, we report CIDEr (blue) and METEOR scores (red)  over the Moments-OVRE test set. }
   \label{fig:abl_qnumber}
   \vspace{-0.5cm} 
\end{figure}

\begin{figure*}[ht]
  \centering
  \includegraphics[width=0.9\textwidth]{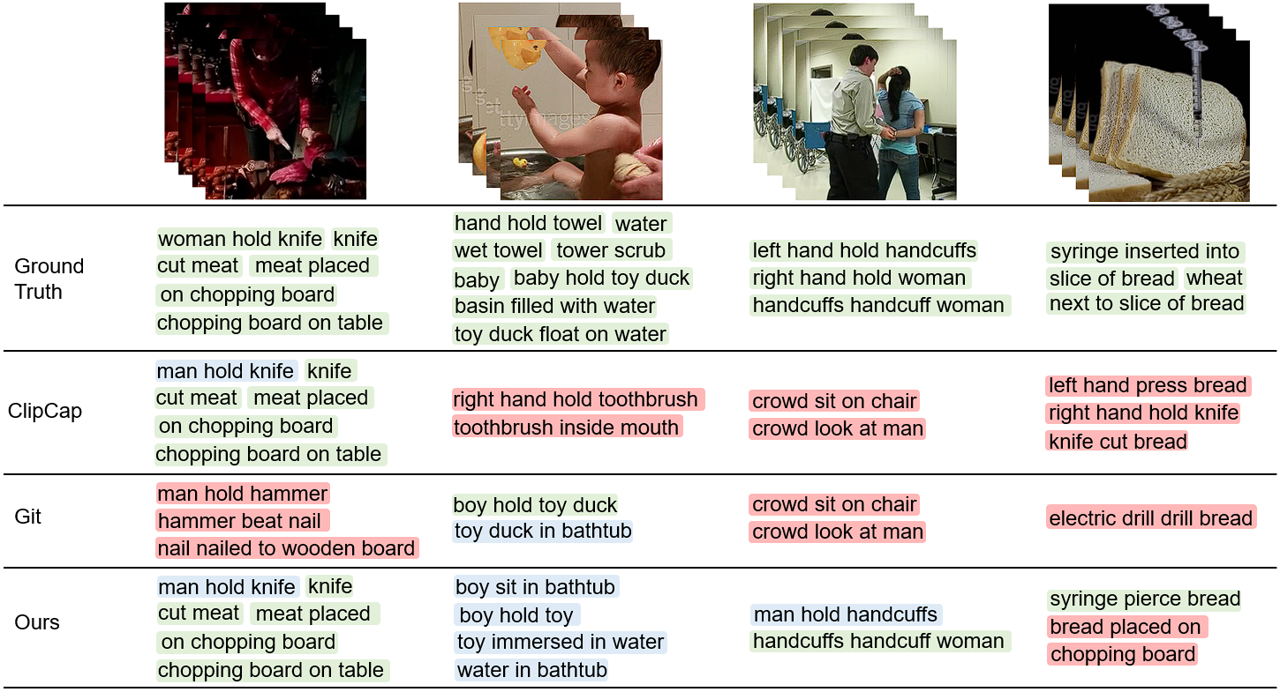}
  \caption{Comparisons of triplets generation across diverse OVRE methods. 
  The illustration highlights accurately described triplets in green, triplets with semantic correlation in blue, and irrelevant triplets in red.
  }
  \label{fig:visualize}
  \vspace{-0.2cm} 
\end{figure*}

\subsection{Ablation Study}
\noindent \textbf{The Impact of Query Numbers. }
We first investigate the impact of query numbers. 
Specifically, we experiment with query number=2, 4, 8, 16, 32, and 64. 
Figure~\ref{fig:abl_qnumber} shows that an extremely small number of queries will yield inferior generation results since the limited queries are insufficient to extract the content in video patches. 
 As the number grows from 2 to 8, the generated results show significant improvement.
 Subsequently, when the number of queries continues to double, the performance of the model gradually becomes saturated. We choose query number=64 as it demonstrated the best performance across all metrics.
 
\noindent \textbf{Exploration of Vision and Language Model Fine-tuning.}
Recent research~\cite{Rasheed2023vificlip} has shown that fully fine-tuned CLIP can effectively bridge the modality gap in the video domain. 
As shown in Table~\ref{tab:abl_ft}, though the trainable parameters increased, fine-tuning the CLIP-ViT can improve the CIDEr score by 8.8.
Additionally, we delve into the outcomes of fine-tuning the text decoder. Experiments reveal that keeping GPT-2's parameters fixed results in a notable reduction in the CIDEr score. We attribute this decline to that GPT-2 is primarily proficient in generating natural language rather than triplets. Therefore, fine-tuning becomes imperative to enhance its capacity for producing rarely seen triplet sequences.

\noindent \textbf{The Effect of Different Visual Features.}
Our experimental investigations involve ablations on different granularity of visual features. Within our proposed framework, we employ patch features extracted from videos as prefixes for the text decoder. Furthermore, we explore two alternative representations as inputs to the attentional pooler:
(\Rmnum{1}) Region features: Following the common VidVRD practice, we extract a sequence of objects and subsequently employ a tracking algorithm to obtain 5 tracklet features per video. These features replace patch features as input to the model. Specifically, we utilize RegionCLIP \cite{zhong2021regionclip} pre-trained from LVIS to crop bounding boxes and seqNMS \cite{han2016seqnms} for object tracking.
(\Rmnum{2}) Frame features: We directly utilize features extracted from individual frames using CLIP, concatenating them to form a representation of frame-level features.
As depicted in Table~\ref{tab:abl_feat}, both frame features and region features exhibit poor performance. Notably, frame features capture the overall visual content of an image but overlook finer details such as objects and relationships. Surprisingly, region features fare even worse compared to frame features. We hypothesize that this is attributed to the limited generalization capability of existing object detectors. The diverse range of object categories complicates their accurate detection within our Moments-OVRE context.


\section{Conclusion}
In this paper, we introduce a new task named OVRE, where the model is required to generate all relationship triplets associated with the video actions. Concurrently, we present the corresponding Moments-OVRE dataset, which encompasses a diverse set of videos along with annotated relationships. We conduct extensive experiments on Moments-OVRE and demonstrated the superiority of our proposed approach over other baseline methods. We hope that our task and dataset will inspire more intricate and generalizable research in the realm of video understanding.

\hspace*{\fill}

\noindent \textbf{Limitations:}
(\Rmnum{1}) This version of Moment-OVRE has currently omitted BBox annotation due to the high cost of annotation. We are committed to progressively enhancing this dataset and intend to introduce BBox annotations in upcoming versions of Moments-OVRE.
(\Rmnum{2}) For extracting action-centric relations, leveraging commonsense among action categories and relations~\cite{yang2018commonsense} or implicit knowledge-driven representation learning methods~\cite{li2023knowledge, li2018deep} have shown promise. We will consider these knowledge-driven methods in future work.

\hspace*{\fill}

\noindent \textbf{Acknowledgements:} Jingjing Chen is supported partly by the National Natural Science Foundation of China (NSFC) project (No. 62072116). Zheng Wang is supported partly by the NSFC project (No. 62302453). Lechao Cheng is supported partly by the NSFC project (No. 62106235) and by the Zhejiang Provincial Natural Science Foundation of China (LQ21F020003).

\bibliography{aaai24}

\end{document}